\documentclass[letterpaper, 10 pt, conference]{ieeeconf}  

\usepackage{soul}
\usepackage{subcaption}
\usepackage{stfloats}
\usepackage{graphicx}
\usepackage{url}

\usepackage{cite}

\usepackage{hyperref}

\IEEEoverridecommandlockouts

\title{\LARGE \bf
A Taxonomy of Conceptual Alignment in Human–Robot Dialogue
}

\author{Shengchen Zhang$^{1}$, Xiaohua Sun$^{2*}$, and Weiwei Guo$^{1*}$
    \thanks{$^{1}$ College of Design and Innovation, Tongji University, Shanghai, China. {\tt\small\{\href{mailto://shengchenzhang@tongji.edu.cn}{shengchenzhang}, \href{mailto://weiweiguo@tongji.edu.cn}{weiweiguo}\}@tongji.edu.cn}}%
    \thanks{$^{2}$ School of Design, Southern University of Science and Technology, Shenzhen, Guangdong, China. {\tt\small \href{mailto://sunxh@sustech.edu.cn}{sunxh@sustech.edu.cn}}}%
    \thanks{$^{*}$ Corresponding authors.}%
    \thanks{This work was supported by the Fundamental Research Funds for the Central Universities from Tongji University and Shanghai Gaofeng Project for University Academic Program Development, the Opening Project of the State Key Laboratory of General Artificial Intelligence (Project No. SKLAGI2025OP14), and the Tongji University national artificial intelligence production-education integration innovation project.}
}

\fnbelowfloat

\begin{document}

\maketitle
\thispagestyle{empty}
\pagestyle{empty}

\begin{abstract}
Successful conversations require speakers to align on the meaning of concepts, a challenging but crucial task for human-robot interaction. Understanding the process of establishing such alignment is hindered by competing interpretations of the term and isolated, unidirectional investigations of its design space. This paper argues for a design-centric understanding of conceptual alignment as a bidirectional and co-constructive process. We introduce a taxonomy that characterizes conceptual alignment dialogues along what triggers its initiation and what level(s) of conceptual understanding it concerns. We further present a dialogue act schema as an operational tool that captures the interactional moves through which alignment is achieved. Together, these contributions provide a structured foundation for analyzing, comparing, and designing conceptual alignment in human-robot interaction.
\end{abstract}
\section{Introduction}

In dialogue, people employ various concepts to refer to complex ideas, reason about categories, and coordinate action.
Enabling robots and other conversational agents to understand and use human concepts has been a long-standing goal, supporting applications such as interactive learning \cite{parkInteractiveAcquisitionFinegrained2023a, kuniyasu_robot_2021, gkatzia_whats_2021}, long-term interaction, and personalization \cite{rane_concept_2024, kirk_benefits_2024}. 
This remains a challenging but crucial problem, not only because concepts must be linked to real-world entities such as objects and places, but also because their meanings are often subjective \cite{sherer_follow_2025}, situational \cite{stolk_conceptual_2016}, and co-constructed in situated interaction \cite{brennan_conceptual_1996}.
In human interaction, such meaning are not fixed in advance, but emerges through interaction in what is commonly referred to as \textit{conceptual alignment} \cite{rane_concept_2024, garrod_saying_1987, braniganLinguisticAlignmentPeople2010, stolk_conceptual_2016}. Through dialogue, speakers iteratively refine, negotiate, and transform their understanding of a concept in order to achieve coordination.

Despite its importance, current HRI research provides only a fragmented understanding of conceptual alignment. Existing work has primarily focused either on 1) enabling robots to acquire specific types of concepts through constrained interaction, 2) studying isolated alignment behaviors such as lexical or referential adaptation, or 3) understanding people's alignment to robots in controlled experimental settings (see Section \ref{alignment-review}). These approaches offer limited insight into how conceptual alignment can be designed and implemented in human-robot dialogue, and how different forms of conceptual knowledge are introduced, negotiated, and reconciled in dialogue. As a result, there is a lack of structured tools for analyzing and designing conceptual alignment in human–robot dialogue.
Moreover, the term ``conceptual alignment'', which is also used in diverse fields including linguistics, cognitive science, artificial intelligence, and human-computer interaction, is often used under differing or unclear interpretations, making it difficult to integrate and compare research findings.

In this paper, we introduce a taxonomy of conceptual alignment dialogue that characterizes alignment along two dimensions, namely what triggers its initiation and what level(s) of conceptual understanding it concerns. We further present a dialogue act schema that captures the interactional moves through which alignment is achieved.
Besides reviewing relevant literature on human and human-robot dialogue, we instantiate this framework through an empirical study of human dialogue in a collaborative concept alignment task. Human-human interaction is used to capture the breadth of alignment behaviors that emerge in natural dialogue, following prior HRI work that derives interaction patterns from human interaction as a basis for robot design. 

This paper makes two main contributions. First, we discuss the interpretations of conceptual alignment for HRI and argue for a design-oriented view of alignment as a deliberate and co-constructive process. Second, we introduce a taxonomy and accompanying dialogue act schema that provide a structured framework for understanding conceptual alignment dialogues. Together, these contributions provide a structured foundation for analyzing, comparing, and designing conceptual alignment in human–robot interaction.
\section{Related Work}
\subsection{Conceptual alignment with robots}
\label{alignment-review}
Equipping robots with human concepts has been a long-standing goal in robotics. Technical methods have been proposed to acquire concepts through dialogue, especially those that are crucial for the robot's function. These include categorical (``stuffed animal'', ``soft object'') \cite{kuniyasu_robot_2021, gkatzia_whats_2021}, perceptible (colors, shapes, angles) \cite{parkInteractiveAcquisitionFinegrained2023a}, actionable (``take'', ``grasp'') \cite{cangelosi_review_2018, bobu_learning_2022}, abstract (numbers) \cite{cangelosi_review_2018}, or referential (``this'', ``that'') \cite{tang-etal-2024-grounding} concepts. 
However, most of these methods frame the process as the robot passively learning concepts from humans, typically with a limited interaction strategy such as a question-asking loop. As a result, the role of dialogue in shaping conceptual understanding is treated as secondary to the learning mechanism itself, and the diversity of interactional strategies for achieving alignment remains underexplored.

Meanwhile, studies on reference and lexical alignment in human-robot dialogue demonstrate that alignment is fundamentally bidirectional and context-dependent, with both humans and robots adapting their terminology based on task goals and interaction dynamics. For instance, people preferred robots to adopt human-generated referential terms rather than imposing their own labeling conventions \cite{kimoto_alignment_2016}. Presenting referential concepts upfront and using full names reduced clarification requests and improved mutual understanding \cite{foster_evaluating_2009}. Matching children’s word choices in storytelling increases acceptance of robot suggestions \cite{calvo-barajas_balancing_2024}. Conversely, humans also actively adapt to robots, consistently aligning their lexical choices \cite{campano_comparative_2014}, mirroring the robot’s conceptual framing \cite{vanlieropConceptualAlignmentReference}, and adjusting their level of abstraction to match the system \cite{cirilloConceptualAlignmentJoint2022}. These complementary findings illustrate that conceptual alignment is not a one-way adaptation but a dynamic, co-constructed process.

Taken together, these lines of work highlight that conceptual alignment in HRI is both bidirectional and design-dependent.
However, there is still limited support for systematically analyzing, comparing, or designing conceptual alignment behaviors in human–robot dialogue. We address this gap by introducing a taxonomy that characterizes both how alignment is achieved and what is being aligned. 

\subsection{HRI taxonomies and patterns}
\label{from-human}
A common approach in HRI research is to develop taxonomies and design patterns that structure complex interaction phenomena into analyzable components. Taxonomies have been used to characterize general HRI \cite{onnaschTaxonomyStructureAnalyze2021a}, mixed-initiative interaction \cite{jiangMixedInitiativeHumanRobotInteraction2015a}, and communicative signals \cite{holthausCommunicativeRobotSignals2023a}. Design patterns have been proposed for robot sociality \cite{kahnDesignPatternsSociality2008}, dialogue \cite{sauppeDesignPatternsExploring2014}, and communication of situational knowledge \cite{10.1145/3411764.3445767}.
Rather than prescribing specific implementations, they provide abstractions that support systematic analysis, comparison across systems, and principled design.
These frameworks can be derived from a combination of prior literature and empirical observations. In particular, human–human interaction has been used as a basis for identifying interactional patterns that can inform HRI design \cite{kahnDesignPatternsSociality2008, sauppeDesignPatternsExploring2014, 10.1145/3411764.3445767}, under the assumption that natural interaction reveals a broader space of strategies than those currently implemented in robotic systems. This approach has been applied in robot sociality and human-robot dialogue, where insights from human interaction have guided the development and evaluation of robot behaviors.
We adopt a similar approach to develop a taxonomy grounded in both prior literature and empirical dialogue data. Our work extends prior work by focusing specifically on conceptual alignment, providing a structured account of how alignment is triggered and unfolds, and what forms of conceptual knowledge are negotiated.
\section{Aligning on ``conceptual alignment'' for HRI}
Before introducing our taxonomy, we clarify how the term ``conceptual alignment'' is interpreted in this work. The term has been studied across linguistics, artificial intelligence, and HRI under differing assumptions, which have led to ambiguity in how alignment is understood and operationalized.
The accounts range from viewing alignment as the \textit{result of an automatic mechanism} where the same concept increasingly means the same thing(s) as a conversation goes on \cite{pickeringMechanisticPsychologyDialogue2004, cirilloConceptualAlignmentJoint2022}, to using it to denote the \textit{deliberate and co-constructive interaction} that ultimately results in mutual conceptual understanding and usage \cite{stolk_conceptual_2016, gastaldonLinguisticAlignmentArtificial2025}. 
Cirillo et al., for example, studied conceptual alignment in the automatic sense, and found that people spontaneously adopted the concepts and level of abstraction used by a social robot \cite{cirilloConceptualAlignmentJoint2022}.
However, a follow-up analysis by Gastaldon and Calignano \cite{gastaldonLinguisticAlignmentArtificial2025} suggests evidence of alignment being ``strategic and effortful'', which seems to lean toward the second interpretation.

For this paper and future HRI research, we argue in favor of the second interpretation.
Although the automatic account may explain how mutual conceptual understanding and usage are achieved among human speakers, how it (and alignment as the \textit{result} of it) can be operationalized and implemented for HRI is under-specified.
Lower-level alignment such as lexical or syntactic alignment concerns linguistic behavior in dialogue only. Therefore, they can be implemented as adopting the same wording or sentence structure in subsequent dialogue generation. The effects of these designs are already studied in HRI, as discussed in Section \ref{alignment-review}.
Yet, alignment at the conceptual level is tied to what internal representation is invoked.
Producing robot dialogue that automatically aligns to human speakers requires (para)linguistic capabilities to link free-form spoken concepts to diverse referents like events, objects, actions, or experiences, and vice versa.
This task is still highly challenging.
As a result, the study and application of automatic conceptual alignment in HRI have remained unidirectional, limited to understanding whether and how people align to robots, which provides an incomplete account for HRI design.
Moreover, implementing such mechanisms likely requires specifying an internal conceptual representation, as well as how to communicate and update it. By doing this, we already drift towards the second interpretation of the term. In short, viewing alignment as the \textit{result} of an automatic process, although having explanatory power for human behaviors, may not translate to the human-robot case.

On the other hand, the interpretation of conceptual alignment as the deliberate and co-constructive \textit{process} by which a mutual understanding is built fits more naturally in the HRI case.
First, by centering alignment interactions and their design, this view makes explicit the mechanisms through which shared meaning is negotiated. This, in turn, enables researchers to analyze, compare, and design alignment behaviors in human–robot dialogue.
Second, by treating alignment as co-constructed, this interpretation also enables a bidirectional view \cite{shen_position_2025} in which both human and robot actively contribute to shaping shared concepts, rather than framing alignment solely as the human adapting to the system. With increasing autonomy and use of generative models in robots, this becomes an important perspective to understand the dynamics of future human-robot interactions.

The taxonomy we propose in the following section builds on this view by characterizing when and how alignment unfolds and what forms of conceptual knowledge are negotiated bi-directionally in interaction. Below, we will first discuss our process of developing the taxonomy, then present its content in detail.

\section{Method}
The taxonomy we present below is developed by combining results from a review of existing literature on dialogue acts related to conceptual alignment with the empirical analysis of a conceptual alignment dialogue corpus we collected.

\subsection{Literature review}
The literature survey focused on models of dialogue processes related to mutual understanding.
Specifically, we searched for work that proposed dialogue acts or models for human, human-agent, and human-robot dialogue in the following areas:
\begin{itemize}
    \item \textbf{Grounding} refers to the process by which dialogue participants build common ground, by actively signaling for (non)understanding and initiating repairs.
    \item \textbf{Meaning negotiation} concerns how meaning for words and concepts is established through negotiation.
    \item \textbf{Argumentation} concerns the process through which dialogue participants present arguments, justification, and attacks to achieve persuasion, thereby establishing a mutually agreed-upon position.
    \item \textbf{Negotiation} dialogues in general complement the relatively limited work in meaning negotiation by providing general dialogue acts that constitute a negotiation.
\end{itemize}

The search started with known papers and iteratively expanded to their citations and references, complemented by database searches in the form of ``robot AND [area terms] AND (model OR "dialogue act" OR "taxonomy")''. Recurring dialogue acts were merged, and the resulting annotation framework is presented in Table \ref{tab:coding-scheme}.

\subsection{Dialogue corpus collection and analysis}
\subsubsection{Data collection}
The dialogue corpus includes 48 spoken dialogues collected from a custom conceptual alignment task. The 48 participants formed 24 pairs, and performed the alignment task and participated in subsequent discussions. Each pair did two versions of the task in random order, resulting in 48 dialogues. The recordings were automatically transcribed and errors manually corrected. In total, we recorded 2677 conversational turns, averaging around 57 turns per discussion.
We opted for human dialogues for two reasons. First, as our goal is to understand \textit{potential} aspects of conceptual alignment, human dialogues provide a larger source of variety. Previous work in HRI has similarly derived interaction patterns from observations of human data \cite{sauppeDesignPatternsExploring2014}. Second, our taxonomy aims to be agnostic to whether it is the human or the robot engaging in alignment, so it becomes necessary to understand and include potential aspects of alignment in human dialogue as well.
 
\begin{figure}[h!]
    \centering
    \includegraphics[width=\linewidth]{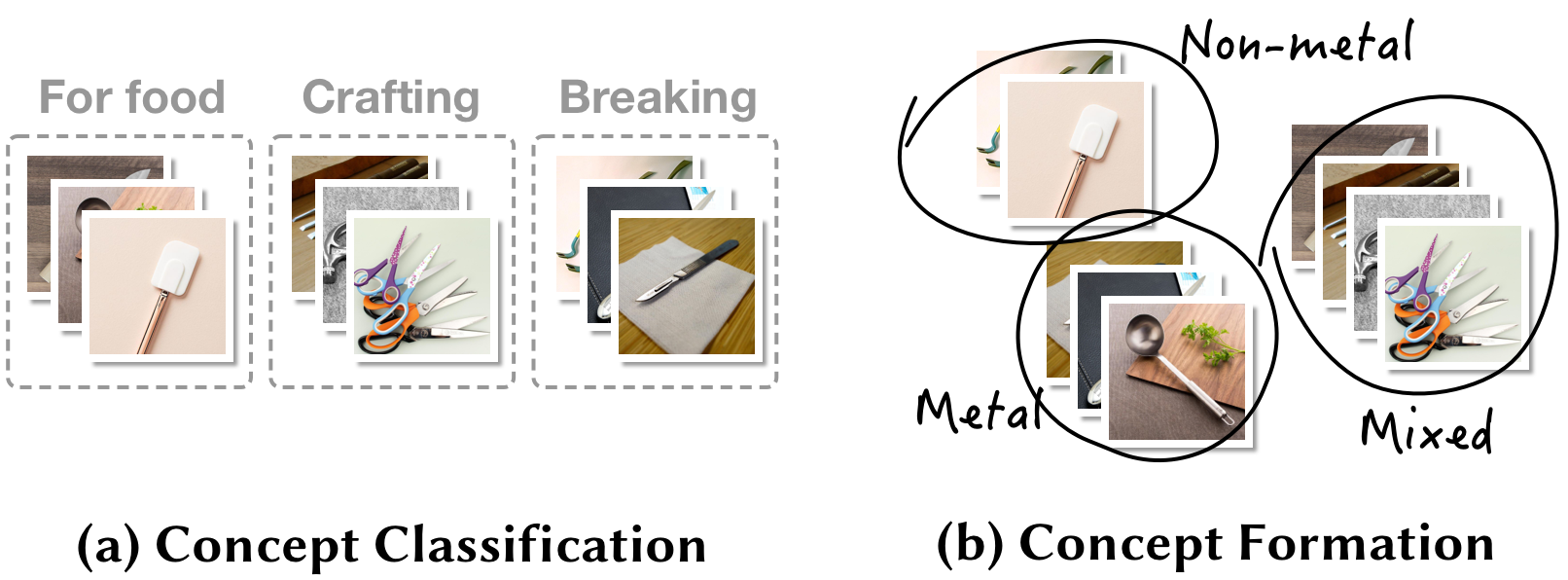}
    \caption{The concept alignment task, where a pair of participants sort images of objects based on their understanding of concepts, and then engaged in discussion. (a) and (b) show two forms of the task.}
    \label{fig:task-design}
\end{figure}

For the alignment task, the participants were given images of everyday objects and sorted the objects according to their conceptual understanding. Then, they engaged in discussion to mitigate differences in understanding. To capture the bidirectional relation between concepts and the real-world phenomena that they represent, we designed two forms of the task: \textbf{classification} and \textbf{formation}. 
The classification task gave participants three predefined concepts (i.e. category names), and asked them to sort the images into these categories according to their understanding (see Figure \ref{fig:task-design}(a)). The formation task asked participants to form concepts (i.e. category names) by themselves, and to sort the images into their self-defined categories (Figure \ref{fig:task-design}(b)).

The minimal design of a conceptual alignment task could simply be asking participants to debate the meaning of provided concepts through discussion. However, image sorting has two advantages. First, this design reflects the features of human-robot dialogue, where the concepts are grounded in real-world entities. Second, in a human-robot interaction context, the objective of conceptual alignment is often to facilitate future interactions/tasks. This operationalization is closer to such an objective than debate alone. Third, it makes explicit the differences in participants' \textit{actual usage of concepts} and therefore may uncover misalignment where oral discussion alone cannot.

\subsubsection{Data analysis}
While the task itself requires participants to sort objects based on their conceptual understanding, we do not consider the entirety of a discussion to be one single dialogue that aligns only on the task-related concepts.
The nature of the task and the ambiguities involved meant that the participants spontaneously initiated sub-dialogues to align on one of the concepts, or concepts that they themselves used during their discussion.
In the following text, when we refer to an \textit{alignment dialogue}, we use the term to denote such a segment of conversational exchange with a consistent topic, aimed at establishing a mutually agreed-upon conceptual understanding for a certain set of concepts.

For analysis, we coded the initiation conditions for each alignment dialogue, as well as individual dialogue acts. For dialogue acts, we first coded them according to the code book developed through our literature review (Table \ref{tab:coding-scheme}), then inductively coded two aspects of each act: 1) how the act was realized in conversation form, and 2) what type of conceptual representation was utilized. These codes were then iteratively refined to form overarching themes.

\section{A taxonomy for conceptual alignment in human-robot dialogue}

In this section, we introduce a taxonomy for conceptual alignment in human-robot dialogue.
Our taxonomy takes a design-centered perspective, viewing
conceptual alignment as a co-constructive process where the HRI design choices shape the alignment outcome together with the interacting human(s).
As such, the taxonomy aims to classify not technical implementations, but aspects to consider when analyzing and designing alignment strategies. The taxonomy contains two levels. At a higher level, it describes a dialogue as a whole, including what \textbf{triggers} its initiation and what \textbf{level(s)} of conceptual understanding it concerns. At a lower level, we further present a set of \textbf{dialogue acts} to aid in analyzing and describing the interaction process in detail.

\begin{figure}[h!]
    \centering
    \includegraphics[width=\linewidth]{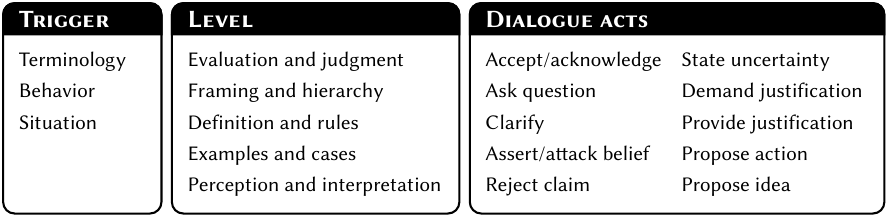}
    \caption{An overview of the proposed taxonomy. Only names are shown here for the dialogue acts. For a detailed list, please refer to Table \ref{tab:coding-scheme}.}
    \label{fig:taxonomy-dialogue}
\end{figure}

\subsection{Triggers for initiation}
First, we define three potential causes for conceptual alignment dialogues to be initiated by an interacting party. These triggers are typically mutually exclusive.

\subsubsection{\textbf{Terminology}}
Alignment dialogues can arise from one speaker using an unknown, vague, ambiguous, or otherwise ``problematic’’ concept. For this, Clark and Schaefer's theory of grounding \cite{clark_contributing_1989} provides an account for when and how such dialogues are initiated, which has long been applied in HRI. When applied to conceptual understanding, the theory predicts that speakers actively look out for evidence that the concepts used have been understood, and initiate repair when the evidence was not found. Listeners also actively provide evidence of ``trouble in understanding'' when a concept could not be interpreted.

\subsubsection{\textbf{Behaviors}}
Another cue for alignment dialogues is later unexpected behaviors that conflicts with the current (perceived) conceptual common ground.
Behavior in this sense refers to both communicative and physical actions, such as those taken to complete a task.
This category differs from the previous by timing, since
many actual and subtle differences in understanding may only be revealed downstream when a previously discussed concept is used in a task or in subsequent dialogues.
When annotating, the context for terminology-triggered dialogues could typically be found in the directly preceding utterance, while behavior-triggered ones refer to context further upstream.

\subsubsection{\textbf{Situation}}
The need for alignment dialogue can also be specified by situational factors, such as the nature of the task at hand or its progress.
While the previous two are reactive responses to misaligned behavior or unclear terminology, alignment dialogues triggered by the situation are proactive.
In our dialogue corpus, participants commonly terminated their discussions by summarizing and confirming their consensus on the definitions or rules for concepts whose meaning they had negotiated.
In this case, the dialogues are initiated by speakers as deemed suitable or helpful, instead of responding to concrete misalignment.

\subsection{Levels of alignment}
An alignment dialogue may utilize one or more types of conceptual representation. Based on existing literature and our analysis of the collected corpus, we distinguish five levels of representations that differ in form, level of abstraction and scope.
Most relevant to this is the work on representation alignment by Bobu et al. \cite{bobu_aligning_2024}. In their framework of representation alignment, they provided a review of common representation types used in robot systems, such as features, embeddings, and graphs. However, this classification remains robot-centric, and is difficult to apply to analyzing human-robot dialogue, where behaviors from both parties need to be taken into account. The levels presented in the following paragraphs are intentionally agnostic to speaker identity, and may be covered by multiple robot representations. For each level, we first present a general definition and description, then discuss corresponding robot-specific representations, and use translated examples from the collected corpus to demonstrate how it applies to concrete dialogue and may translate to the human-robot case. The example dialogues were translated from Chinese text by the first author.

\subsubsection{\textbf{Perception and its interpretation}}
At this level, the speakers aim to build \textit{mutual interpretation of perception}, i.e., how perceptual features are mapped to meaningful, agreed-upon categories. 
In robotic systems, this level typically corresponds to representations of sensor perception, where things, such as objects, events, and actions, are described through values that map to colors, shapes, parameters, affordances, etc. 
They could be represented as feature-specific models parameters, embeddings, or attributes and properties in knowledge-based systems. 
Together, they can then be used to assign conceptual membership.
However, alignment at this level is not merely about mapping percepts, but about agreeing on their \textit{interpretation} in context.
Depending on robot capabilities, task goals or user expectations, the same percept may require different mappings. 
In our dialogue corpus, we observed that people need to align on this level despite having similar perceptual bases and general knowledge due to the current context. Take the following dialogue:

\begin{quote}
        \begin{description}
      \item[\textbf{P46:}] Or if we go with "singular objects", like, [...] the ones that are inseparable, like that tomato. And [...]
      \item[] \textit{[Unrelated turns omitted for brevity.]}
      \item[\textbf{P45:}] The... um. [The things you call] ``inseparable''... Do you mean it's a single solid shape? Because for example the tomato, you can take the [stem] on top ... [and separate it]
    \end{description}
\end{quote}

The need for alignment on ``inseparable'' arises here because a tomato can be interpreted either as a single object or as two separate objects (the fruit and the stem). In a human-robot scenario, this would correspond to aligning on the granularity of open-domain segmentation when processing a visual scene. Alignment at this level matters because mismatches in feature interpretation could further propagate to downstream processing, leading to systematic disagreement about category membership, user intent, or task structure.

\subsubsection{\textbf{Examples and cases}}
The next matter of concern is the conceptual membership of certain examples or cases. Alignment dialogues on this level concern the inclusion or exclusion of specific instances. Speakers negotiate conceptual boundaries through reference to current or hypothetical examples, often using cases to refine or challenge prior conceptions.
This form of alignment is consistent with many exemplar-based concept learning techniques, including most reviewed in the related work section.
Dialogue for this level of alignment is prevalent in the corpus. The following dialogue shows one challenging the other's conception using both real and hypothetical examples:

\begin{quote}
        \begin{description}
      \item[\textbf{P39:}] Say, the water bottle on your side. [...] When you have period cramps, and you use a hot water bottle, it's satisfying your psychological needs.
      \item[\textbf{P40:}] Hah, That's true... Really, like, if [one] categorizes them by their function, then it actually serves multiple [functions].
      \item[\textbf{P39:}] Right. [...] And if like, for pills. [...] If it [treats] psychological conditions, then it seems okay to say that it satisfies psychological needs.
    \end{description}
\end{quote}

\subsubsection{\textbf{Definition and rules}}
While the previous level deals with concrete features and instances, concepts can also be discussed in terms of rules and definitions.
Alignment on this level is therefore focused on devising an agreed-upon definition or set of rules that generalizes to future cases.
For implementation in robotic systems, examples that represent concepts at such a level are ontology-based methods (e.g., see \cite{manzoor_ontology-based_2021} and \cite{olivares-alarcosReviewComparisonOntologybased2019a} for a review), where concepts are represented as standalone classes with fixed definition and predefined properties, or logical predicates that derives class membership based a combination of subclass relations and properties.
Participants in our study commonly spoke at this level at the beginning or finalizing stages of their discussion. The example below happened at the beginning of their discussion:

\begin{quote}
    \begin{description}
      \item[\textbf{P1:}] [...] Biological things are... Like, it's still growing, and nobody touched it? [...]
      \item[\textbf{P2:}] Is this the definition of biological (said as \textit{shēng wù} in Chinese)?
      \item[\textbf{P1:}] Like... living (literal meaning of \textit{shēng}) things (literal meaning of \textit{wù}).
    \end{description}
\end{quote}

\subsubsection{\textbf{Framing and hierarchy}}
The previous levels mostly concern specific concepts and their members. Yet, what concepts are used to express or characterize a problem space, and the relation between these concepts, is also a concern for alignment.
This level surfaces how different people or robot systems structure the conceptual space itself. By making this level explicit in the taxonomy, it enables analysis of whether alignment operates within a fixed ontology or allows restructuring through interaction.
For representations to support this level of alignment, mechanisms to update not only the represented content but also the representation itself are needed. This may include continuous learning techniques that directly update a model's architecture, or dynamically updating the representation structure in knowledge-based methods.
Below is an example from the corpus that attempts to align on the level of framing and hierarchy: 

\begin{quote}
    \begin{description}
      \item[\textbf{P13:}] [...] So you are sorting things based on how it is used? I was sorting them by material.
      \item[\textbf{P14:}] [...] I now suddenly realize that I was sorting things according to a progression from production to consumption. Like some things are not yet consumer products, so they're one category. Then consumer products are another one.
      \item[\textbf{P13:}] Mm. Then, if I were to challenge you... [...] Okay so maybe [what you just said] were all consumer products. It's just one group [you buy and] use in the process of making something. The other ones you can directly use in your life.
    \end{description}
\end{quote}

Here, P14 is not explaining one concept but the relations between two concepts. Similarly, P13 challenges P14 in the end with an alternative understanding, which reframes how the same two groups of objects are conceptualized under a hierarchy. In human-robot dialogue, this could likely apply to, for example, building ad-hoc ontological hierarchies for a new or personalized domain, or negotiating a process among many possible ways to satisfy a user need.

\subsubsection{\textbf{Evaluation and judgment}}
Finally, speakers impose their evaluation and judgment onto their current alignment status, as well as the concepts, framing, and hierarchy used in a task. Alignment at this level may be considered ``meta-level'' \cite{pickeringMechanisticPsychologyDialogue2004}, as what is being calibrated is not conceptual understanding itself, but the speakers' evaluations of their current alignment. There is no representation specific to conceptual understanding in robot implementations for this level to the best of our knowledge. However, we'd like to note that general human-robot dialogue frameworks based on argumentation (e.g., \cite{sklar_argumentation-based_2015}) offer structures that represent speaker positions and their current agreement.

In the dialogue corpus, we found people to periodically provide both assessments of their current alignment and judgment of whether certain conceptions are effective for the task at hand.

\begin{quote}
    \begin{description}
      \item[\textbf{P20:}] I also thought about putting all tools in one place. But I felt a lot could be in there... but also not exactly. So I sort of expanded its definition and put many things in there.
      \item[\textbf{P19:}] Mm. That's fine. I can understand it.
      \item[\textbf{P20:}] Then I also feel like I can understand both these sorting methods. And I can agree.
      \item[] \textit{[Unrelated turns omitted for brevity.]}
      \item[\textbf{P19:}] I feel like we are kind of there?   
      \item[\textbf{P20:}] [...] I think we have a consensus.
    \end{description}
\end{quote}

The first portion of the dialogue is not about \textit{what} conception to adopt. Instead, the speakers discuss whether they are useful to the task and understandable by both. The second portion, similarly, saw the two speakers checking whether they both feel like they had a consensus.

In reality, the levels presented above were often intertwined, with speakers shifting between them in a dialogue or across dialogues. When annotating a transcript, these levels typically apply to individual or continuous spans of dialogue acts, instead of an entire dialogue as a whole.
The value of this decomposition lies in providing a structured lens to analyze and compare conceptual alignment across different systems and interaction settings. By making explicit the representational and interactional assumptions underlying alignment, the taxonomy allows HRI researchers and designers to examine which aspects of conceptual understanding are supported, omitted, or mismatched in a given robot system or a specific dialogue.

\subsection{Dialogue acts}
\label{dialogue-acts}

\begin{table*}[b!]
    \caption{Common dialogue acts for conceptual alignment dialogue, operationalized as a codebook for annotation.}
    \label{tab:coding-scheme}
    \begin{tabular}{lp{2.4in}p{1.9in}p{1.1in}}
        \hline
        Code  & Definition & Example & References \\
        \hline
        Accept/acknowledge
        & Show that a statement is understood or agreed. 
        & ``Yeah, it makes sense.''
        & \cite{brennanGroundingProblemConversations}\cite{santosDialogueProtocolSupport2016}\cite{prakken_models_2009}\cite{sklar_argumentation-based_2015}\cite{dillenbourg_negotiation_1996}\cite{mohapatra_conversational_2023} \\
        
        Ask question
        & Ask a question with the intention of getting new information.
        & ``Can you define `mechanical'?''
        & \cite{walton_commitment_1995}\cite{prakken_models_2009}\cite{mcburney_locutions_2005}\cite{sklar_argumentation-based_2015}\\

        Clarify
        & Ask a question with the intention of getting verification or clarification.
        & ``Are you referring to the blueberries?''
        & \cite{brennanGroundingProblemConversations, mohapatra_conversational_2023}\\
        
        Assert belief 
        & State a fact or opinion that the speaker holds. 
        & ``I think [this object] is mechanical.''
        & \cite{sklar_argumentation-based_2015}\cite{ayoobi_argumentation-based_2022}\cite{prakken_models_2009}\cite{mcburney_locutions_2005}\\

        Attack belief 
        & Express disagreement with a particular belief of the other speaker. 
        & ``A fire alarm isn't an `office supply'.''
        & \cite{sklar_argumentation-based_2015}\cite{ayoobi_argumentation-based_2022}
        \\

        Reject Claim
        & Express non-acceptance for the other person's justification or proposal. 
        & ``No need to do that.''
        & \cite{santosDialogueProtocolSupport2016}\cite{sklar_argumentation-based_2015}\cite{dillenbourg_negotiation_1996}\\

        State uncertainty 
        & Express uncertainty regarding a fact or opinion. 
        & ``I don't know what it is.''
        & \cite{sklar_argumentation-based_2015}
        \\
        
        Demand justification
        & Demand justification for a statement or action.
        & ``Why did you put it in this category?''
        & \cite{walton_commitment_1995}\cite{prakken_models_2009}\cite{mcburney_locutions_2005}\cite{sklar_argumentation-based_2015}\\
        
        Provide justification  
        & Provide a statement that is believed to justify a statement or action. 
        & ``Because it has nothing to do with analog signals.''
        & \cite{santosDialogueProtocolSupport2016}\cite{prakken_models_2009}\cite{mcburney_locutions_2005}\cite{sklar_argumentation-based_2015}\\
        
        Propose action
        & Propose a (joint) action to be taken.
        & ``Let's discuss the other concepts.''
        & \cite{walton_commitment_1995}\cite{santosDialogueProtocolSupport2016}\cite{walkerDATEDialogueAct2001}\cite{dillenbourg_negotiation_1996}\\
        
        Propose idea
        & State a fact or opinion to be possible without implying truthfulness.
        & ``...Maybe it's also for entertainment.''
        & \cite{walton_commitment_1995}\cite{sklar_argumentation-based_2015}\\
        
        \hline
    \end{tabular}
\end{table*}

The two dimensions outlined above provide a high-level characterization of conceptual alignment dialogues in terms of what is being aligned and under what conditions it is initiated. However, to analyze how alignment unfolds moment-to-moment in interaction, a more fine-grained and operational representation is required. We present a set of dialogue acts that capture the interactional moves through which participants negotiate conceptual meaning. The act set is informed by our literature review, and further instantiated through the analysis of our corpus. 
Table \ref{tab:coding-scheme} shows an overview of the dialogue acts, their definitions, as well as example quotations and references. Below, we describe two notable insights into how some of the acts are realized in conversation that could use special care in annotation, or could potentially inform HRI design.

\subsubsection{Means of justification}
Although treated as one single act in the literature, we further derived three sub-codes under the \textit{provide justification} act through the analysis. Namely, participants justified their conceptual choices and understanding through not only \textbf{logical}, but also \textbf{practical} and \textbf{experiential} arguments.
Logical arguments are typically associated with definitions or rules and their applicability in certain cases, and most closely resembles ``justifications'' as defined in the literature. However, people also justified their conceptual understanding by practicality, stating that definitions can be tweaked away from common understanding to fulfill task requirements. Others justified their understanding by personal experience, such as how they personally used some objects in the task. These phenomena highlight that interpretations and application of a concept can differ based on the task and the interacting person, which need to be considered when implementing conceptual alignment in HRI.

\subsubsection{Questions vs. requests for justification}
The literature we surveyed typically made distinctions between informational questions and demands for justification. However, in practice we found it difficult at times to distinguish between the two when analyzing our corpus. Take the following two utterances: 1) ``What's your definition of `artificial' things?'' and 2) ``Why is [it] artificial?'' While both are questions, the first one was annotated as an informational question, while the latter as a demand for justification. The distinction we defined for our codebook was that demands for justifications are always directed either at a previous statement or action, while informational questions are not. We decided not to merge the two codes because informational questions were often categorized as a part of grounding, while demands for justification a part of argumentation dialogues. Keeping the two codes could help distinguish between the two types of dialogues in future analyses.
\section{Discussion and conclusion}

\paragraph{Using the taxonomy}
The taxonomy can be applied as an analytical tool for studying conceptual alignment in both human–human and human–robot interaction. Researchers may first identify alignment dialogues based on linguistic or behavioral cues indicating potential misalignment, as discussed in the taxonomy as ``triggers'', and then characterize these episodes along the levels of conceptual representation aligned. The dialogue act schema can further be used to annotate interactional moves, enabling fine-grained analysis of alignment strategies.
The taxonomy therefore supports comparative evaluation across systems. By making explicit which levels of alignment are addressed and which alignment acts were employed, the framework allows researchers to examine differences in alignment behavior and relate them to task outcomes or user perceptions. In this way, the taxonomy provides a basis for accumulating knowledge about conceptual alignment across studies.

\paragraph{Implications for HRI and dialogue design}

Rather than prescribing specific interaction strategies, the taxonomy highlights a design space for conceptual alignment in HRI. Different systems may support alignment at different levels (e.g., examples vs. conceptual framing), be initiated differently (e.g., reacting to explicit clarification vs. proactively addressing unexpected behaviors), and employ different dialogue actions to produce different behaviors. By making these dimensions explicit, the framework enables reasoning about which forms of alignment are supported or omitted in a given system, and how these choices may impact interaction. This is particularly relevant for systems that rely on recent generative models, where alignment behaviors are often implicit and difficult to control.
Conceptual alignment also connects to several core concerns in HRI. For example, transparency and explainability concern whether a robot can make its internal representations understandable to users, while value alignment requires abstract preferences and goals be interpreted as intended. In dialogue-based approaches, both cases could benefit from negotiating conceptual meaning in context, by applying conceptual alignment as a lens to understand and structure this process.

\paragraph{Limitations}
First, the taxonomy is instantiated through a representative but still controlled concept alignment task. This allowed us to systematically observe rich alignment behaviors, but in-the-wild interaction and other domains (e.g., long-term interaction or open-ended social dialogue) may involve additional alignment dynamics not fully represented here.
Second, our analysis is based on human dialogues, which did not include robotic constraints such as differences in perception, embodiment, and capabilities. However, this enabled us to capture a broad range of naturally occurring alignment strategies without constraining interaction with current system capabilities, following prior HRI work that derives interaction patterns from human interaction (Section \ref{from-human}). Future work should evaluate this taxonomy on HRI dialogue corpora or situated tasks.

Overall, this paper positions conceptual alignment as a central mechanism in human-robot representation alignment and communication, connecting low-level perception with high-level representations through co-constructing conceptual meaning. By providing a structured framework for analyzing and designing alignment, we aim to support designerly efforts towards more transparent, adaptive, and collaborative human–robot interaction and dialogue.

\bibliographystyle{IEEEtran}
\bibliography{references}

@inproceedings{gkatzia_whats_2021,
	series = {{ACM}/{IEEE} {International} {Conference} on {Human}-{Robot} {Interaction}},
    booktitle = {Companion of the 2021 ACM/IEEE International Conference on Human-Robot Interaction},
    address = {Boulder, CO, USA},
	title = {"{What}'s this?" {Comparing} {Active} learning {Strategies} for {Concept} {Acquisition} in {HRI}},
	isbn = {2167-2121},
	url = {https://www.scopus.com/inward/record.uri?eid=2-s2.0-85102767650&doi=10.1145%2f3434074.3447160&partnerID=40&md5=a860b5d79cac3847285a840f31c70d4e},
	doi = {10.1145/3434074.3447160},
	language = {English},
	publisher = {IEEE Computer Society},
	author = {Gkatzia, Dimitra and Belvedere, Francesco},
	year = {2021},
	pages = {205--209},
}

@article{kuniyasu_robot_2021,
	title = {Robot {Concept} {Acquisition} {Based} on {Interaction} {Between} {Probabilistic} and {Deep} {Generative} {Models}},
	volume = {3},
	issn = {2624-9898},
    numpages = {14},
	url = {https://www.scopus.com/inward/record.uri?eid=2-s2.0-85117894920&doi=10.3389%2ffcomp.2021.618069&partnerID=40&md5=eff3ee9b0230e939c913f630bf6eb727},
	doi = {10.3389/fcomp.2021.618069},
	language = {English},
	journal = {Frontiers in Computer Science},
	author = {Kuniyasu, Ryo and Nakamura, Tomoaki and Taniguchi, Tadahiro and Nagai, Takayuki},
	month = sep,
	year = {2021},
}

@article{manzoor_ontology-based_2021,
	title = {Ontology-{Based} {Knowledge} {Representation} in {Robotic} {Systems}: {A} {Survey} {Oriented} toward {Applications}},
	volume = {11},
	issn = {2076-3417},
	shorttitle = {Ontology-{Based} {Knowledge} {Representation} in {Robotic} {Systems}},
	url = {https://www.mdpi.com/2076-3417/11/10/4324},
	doi = {10.3390/app11104324},
	language = {en},
	number = {10},
	urldate = {2023-01-15},
	journal = {Applied Sciences},
	author = {Manzoor, Sumaira and Rocha, Yuri Goncalves and Joo, Sung-Hyeon and Bae, Sang-Hyeon and Kim, Eun-Jin and Joo, Kyeong-Jin and Kuc, Tae-Yong},
	month = may,
	year = {2021},
	pages = {4324},
}

@inproceedings{parkInteractiveAcquisitionFinegrained2023a,
  title = {Interactive {{Acquisition}} of {{Fine-grained Visual Concepts}} by {{Exploiting Semantics}} of {{Generic Characterizations}} in {{Discourse}}},
  booktitle = {Proceedings of the 15th {{International Conference}} on {{Computational Semantics}}},
  author = {Park, Jonghyuk and Lascarides, Alex and Ramamoorthy, Subramanian},
  editor = {Amblard, Maxime and Breitholtz, Ellen},
  year = {2023},
  month = jun,
  pages = {318--331},
  publisher = {Association for Computational Linguistics},
  address = {Nancy, France},
  urldate = {2025-01-24}
}

@inproceedings{dillenbourg_negotiation_1996,
	title = {Negotiation spaces in human-computer collaborative learning},
	author = {Dillenbourg, Pierre and Baker, Michael},
	month = jul,
	year = {1996},
	pages = {187--206},
	booktitle = {Actes du colloque {COOP}'96},
	address = {Juan-les-Pins, France},
    publisher = {INRIA},
}

@article{ayoobi_argumentation-based_2022,
	title = {Argumentation-{Based} {Online} {Incremental} {Learning}},
	volume = {19},
	issn = {1545-5955, 1558-3783},
	url = {https://ieeexplore.ieee.org/document/9592832/},
	doi = {10.1109/TASE.2021.3120837},
	language = {en-US},
	number = {4},
	urldate = {2023-03-20},
	journal = {IEEE Transactions on Automation Science and Engineering},
	author = {Ayoobi, Hamed and Cao, Ming and Verbrugge, Rineke and Verheij, Bart},
	month = oct,
	year = {2022},
	pages = {3419--3433},
}

@inproceedings{campano_comparative_2014,
	title = {Comparative analysis of verbal alignment in human-human and human-agent interactions},
    booktitle = {Proceedings of the Ninth International Conference on Language Resources and Evaluation (LREC'14)},
    publisher = {European Language Resources Association (ELRA)},
    address = {Reykjavik, Iceland},
	url = {https://www.semanticscholar.org/paper/Comparative-analysis-of-verbal-alignment-in-and-Campano-Durand/3abdcc2d0d0be072cf46b0b426061b041c74a658},
	urldate = {2024-09-13},
	author = {Campano, Sabrina and Durand, Jessica and Clavel, C.},
	month = may,
	year = {2014},
	pages = {4415-4422},
}

@inproceedings{santosDialogueProtocolSupport2016,
  title = {A {{Dialogue Protocol}} to {{Support Meaning Negotiation}} ({{Extended Abstract}})},
  numpages = {2},
  booktitle = {Proceedings of the 15th {{International Conference}} on {{Autonomous Agents}} and {{Multiagent Systems}}},
  author = {Santos, Gabrielle and Tamma, Valentina and Payne, Terry R and Grasso, Floriana and Santos, G S J and Tamma, V and Payne, T R},
  year = {2016-05-09/2016-05-13},
  publisher = {{International Foundation for Autonomous Agents and Multiagent Systems}},
  address = {Singapore},
  langid = {american}
}

@inproceedings{tang-etal-2024-grounding,
    title = "Grounding Language in Multi-Perspective Referential Communication",
    author = "Tang, Zineng  and
      Mao, Lingjun  and
      Suhr, Alane",
    editor = "Al-Onaizan, Yaser  and
      Bansal, Mohit  and
      Chen, Yun-Nung",
    booktitle = "Proceedings of the 2024 Conference on Empirical Methods in Natural Language Processing",
    month = nov,
    year = "2024",
    address = "Miami, Florida, USA",
    publisher = "Association for Computational Linguistics",
    url = "https://aclanthology.org/2024.emnlp-main.1100/",
    doi = "10.18653/v1/2024.emnlp-main.1100",
    pages = "19727--19741"
}

@article{sklar_argumentation-based_2015,
	title = {Argumentation-based dialogue games for shared control in human-robot systems},
	volume = {4},
	doi = {10.5898/JHRI.4.3.Sklar},
	language = {en-US},
	number = {3},
	journal = {Journal of Human-Robot Interaction},
	author = {Sklar, Elizabeth I. and Azhar, M. Q.},
	month = dec,
	year = {2015},
	pages = {120--148},
}

@inproceedings{mcburney_locutions_2005,
	address = {Berlin, Heidelberg},
	series = {Lecture {Notes} in {Computer} {Science}},
	title = {Locutions for {Argumentation} in {Agent} {Interaction} {Protocols}},
	isbn = {978-3-540-32258-0},
	doi = {10.1007/978-3-540-32258-0_14},
	language = {en},
	booktitle = {Agent {Communication}},
	publisher = {Springer},
	author = {McBurney, Peter and Parsons, Simon},
	editor = {van Eijk, Rogier M. and Huget, Marc-Philippe and Dignum, Frank},
	year = {2005},
	pages = {209--225},
}

@inproceedings{walkerDATEDialogueAct2001,
  title = {{{DATE}}: {{A Dialogue Act Tagging Scheme}} for {{Evaluation}} of {{Spoken Dialogue Systems}}},
  shorttitle = {{{DATE}}},
  booktitle = {Proceedings of the {{First International Conference}} on {{Human Language Technology Research}}},
  author = {Walker, Marilyn and Passonneau, Rebecca},
  year = {2001},
  urldate = {2025-01-24},
  address = {},
  publisher = {},
  pages = {1--8}
}

@book{walton_commitment_1995,
	address = {Albany},
	series = {{SUNY} series in logic and language},
	title = {Commitment in dialogue: basic concepts of interpersonal reasoning},
	isbn = {978-0-7914-2585-5 978-0-7914-2586-2},
	shorttitle = {Commitment in dialogue},
	language = {en-US},
	publisher = {State University of New York Press},
	author = {Walton, Douglas N. and Krabbe, E. C. W.},
	year = {1995},
}

@incollection{prakken_models_2009,
	address = {Boston, MA},
	title = {Models of {Persuasion} {Dialogue}},
	isbn = {978-0-387-98197-0},
	url = {https://doi.org/10.1007/978-0-387-98197-0_14},
	language = {en},
	urldate = {2023-07-28},
	booktitle = {Argumentation in {Artificial} {Intelligence}},
	publisher = {Springer US},
	author = {Prakken, Henry},
	editor = {Simari, Guillermo and Rahwan, Iyad},
	year = {2009},
	doi = {10.1007/978-0-387-98197-0_14},
	pages = {281--300},
}

@inproceedings{kimoto_alignment_2016,
	address = {New York, NY, USA},
	series = {{HAI} '16},
	title = {Alignment {Approach} {Comparison} between {Implicit} and {Explicit} {Suggestions} in {Object} {Reference} {Conversations}},
	isbn = {978-1-4503-4508-8},
	url = {https://dl.acm.org/doi/10.1145/2974804.2974814},
	doi = {10.1145/2974804.2974814},
	urldate = {2024-08-22},
	booktitle = {Proceedings of the {Fourth} {International} {Conference} on {Human} {Agent} {Interaction}},
	publisher = {Association for Computing Machinery},
	author = {Kimoto, Mitsuhiko and Iio, Takamasa and Shiomi, Masahiro and Tanev, Ivan and Shimohara, Katsunori and Hagita, Norihiro},
	month = oct,
	year = {2016},
	pages = {193--200},
}

@inproceedings{bobu_aligning_2024,
	address = {New York, NY, USA},
	series = {{HRI} '24},
	title = {Aligning {Human} and {Robot} {Representations}},
	isbn = {9798400703225},
	url = {https://dl.acm.org/doi/10.1145/3610977.3634987},
	doi = {10.1145/3610977.3634987},
	urldate = {2024-08-22},
	booktitle = {Proceedings of the 2024 {ACM}/{IEEE} {International} {Conference} on {Human}-{Robot} {Interaction}},
	publisher = {Association for Computing Machinery},
	author = {Bobu, Andreea and Peng, Andi and Agrawal, Pulkit and Shah, Julie A and Dragan, Anca D.},
	month = mar,
	year = {2024},
	pages = {42--54},
}

@article{calvo-barajas_balancing_2024,
	title = {Balancing {Human} {Likeness} in {Social} {Robots}: {Impact} on {Children}’s {Lexical} {Alignment} and {Self}-disclosure for {Trust} {Assessment}},
	shorttitle = {Balancing {Human} {Likeness} in {Social} {Robots}},
	url = {https://dl.acm.org/doi/10.1145/3659062},
	doi = {10.1145/3659062},
	urldate = {2024-08-12},
	journal = {J. Hum.-Robot Interact.},
	author = {Calvo-Barajas, Natalia and Akkuzu, Anastasia and Castellano, Ginevra},
	month = may,
	year = {2024},
	note = {Just Accepted},
	volume = {13},
	number = {4},
	pages = {1--27},
}

@article{braniganLinguisticAlignmentPeople2010,
  title = {Linguistic Alignment between People and Computers},
  author = {Branigan, Holly P. and Pickering, Martin J. and Pearson, Jamie and McLean, Janet F.},
  year = {2010},
  month = sep,
  journal = {Journal of Pragmatics},
  series = {How People Talk to {{Robots}} and {{Computers}}},
  volume = {42},
  number = {9},
  pages = {2355--2368},
  issn = {0378-2166},
  doi = {10.1016/j.pragma.2009.12.012},
  urldate = {2025-01-21}
}

@article{cirilloConceptualAlignmentJoint2022,
  title = {Conceptual Alignment in a Joint Picture-Naming Task Performed with a Social Robot},
  author = {Cirillo, Giusy and Runnqvist, Elin and Strijkers, Kristof and Nguyen, No{\"e}l and Baus, Cristina},
  year = {2022},
  month = oct,
  journal = {Cognition},
  volume = {227},
  pages = {105213},
  issn = {0010-0277},
  doi = {10.1016/j.cognition.2022.105213},
  urldate = {2025-01-21}
}

@article{vanlieropConceptualAlignmentReference,
  title = {Conceptual Alignment in Reference with Artificial and Human Dialogue Partners},
  author = {{van Lierop}, Koen and Goudbeek, Martijn and Krahmer, Emiel},
  year = {2012},
  journal = {Proceedings of the Annual Meeting of the Cognitive Science Society},
  volume = {34},
  number = {34},
  pages = {1066--1071},
  issn = {1069-7977},
  langid = {english}
}

@incollection{brennanGroundingProblemConversations,
  title = {The {{Grounding Problem}} in {{Conversations With}} and {{Through Computers}}},
  booktitle = {Social and Cognitive Psychological Approaches to Interpersonal Communication},
  author = {Brennan, Susan E},
  year = {1998},
  pages = {201--225},
  address = {Hillsdale, NJ},
  langid = {english},
  publisher = {Psychology Press}
}

@misc{shen_position_2025,
	title = {Position: {Towards} {Bidirectional} {Human}-{AI} {Alignment}},
	shorttitle = {Position},
	url = {http://arxiv.org/abs/2406.09264},
	doi = {10.48550/arXiv.2406.09264},
	language = {en},
	urldate = {2026-01-19},
	publisher = {arXiv},
	author = {Shen, Hua and Knearem, Tiffany and Ghosh, Reshmi and Alkiek, Kenan and Krishna, Kundan and Liu, Yachuan and Ma, Ziqiao and Petridis, Savvas and Peng, Yi-Hao and Qiwei, Li and Rakshit, Sushrita and Si, Chenglei and Xie, Yutong and Bigham, Jeffrey P. and Bentley, Frank and Chai, Joyce and Lipton, Zachary and Mei, Qiaozhu and Mihalcea, Rada and Terry, Michael and Yang, Diyi and Morris, Meredith Ringel and Resnick, Paul and Jurgens, David},
	month = sep,
	year = {2025},
}

@article{clark_contributing_1989,
	title = {Contributing to {Discourse}},
	volume = {13},
	issn = {1551-6709},
	url = {https://onlinelibrary.wiley.com/doi/abs/10.1207/s15516709cog1302_7},
	doi = {10.1207/s15516709cog1302_7},
	language = {en},
	number = {2},
	urldate = {2025-12-15},
	journal = {Cognitive Science},
	author = {Clark, Herbert H. and Schaefer, Edward F.},
	year = {1989},
	pages = {259--294},
}

@article{stolk_conceptual_2016,
	title = {Conceptual {Alignment}: {How} {Brains} {Achieve} {Mutual} {Understanding}},
	volume = {20},
	issn = {1364-6613, 1879-307X},
	shorttitle = {Conceptual {Alignment}},
	url = {https://www.cell.com/trends/cognitive-sciences/abstract/S1364-6613(15)00286-7},
	doi = {10.1016/j.tics.2015.11.007},
	language = {English},
	number = {3},
	urldate = {2026-01-07},
	journal = {Trends in Cognitive Sciences},
	publisher = {Elsevier},
	author = {Stolk, Arjen and Verhagen, Lennart and Toni, Ivan},
	month = mar,
	year = {2016},
	pages = {180--191},
}

@article{kirk_benefits_2024,
	title = {The benefits, risks and bounds of personalizing the alignment of large language models to individuals},
	volume = {6},
	copyright = {2024 Springer Nature Limited},
	issn = {2522-5839},
	url = {https://www.nature.com/articles/s42256-024-00820-y},
	doi = {10.1038/s42256-024-00820-y},
	language = {en},
	number = {4},
	urldate = {2024-08-12},
	journal = {Nature Machine Intelligence},
	author = {Kirk, Hannah Rose and Vidgen, Bertie and Röttger, Paul and Hale, Scott A.},
	month = apr,
	year = {2024},
	pages = {383--392},
}

@misc{rane_concept_2024,
	title = {Concept {Alignment}},
	url = {http://arxiv.org/abs/2401.08672},
	doi = {10.48550/arXiv.2401.08672},
	language = {en-US},
	urldate = {2024-08-11},
	publisher = {arXiv},
	author = {Rane, Sunayana and Bruna, Polyphony J. and Sucholutsky, Ilia and Kello, Christopher and Griffiths, Thomas L.},
	month = jan,
	year = {2024},
}

@article{garrod_saying_1987,
	title = {Saying what you mean in dialogue: {A} study in conceptual and semantic co-ordination},
	volume = {27},
	issn = {0010-0277},
	shorttitle = {Saying what you mean in dialogue},
	url = {https://www.sciencedirect.com/science/article/pii/0010027787900187},
	doi = {10.1016/0010-0277(87)90018-7},
	language = {en-US},
	number = {2},
	urldate = {2024-08-22},
	journal = {Cognition},
	author = {Garrod, Simon and Anderson, Anthony},
	month = nov,
	year = {1987},
	pages = {181--218},
}

@inproceedings{foster_evaluating_2009,
	title = {Evaluating {Description} and {Reference} {Strategies} in a {Cooperative} {Human}-{Robot} {Dialogue} {System}},
	language = {en},
	booktitle = {Proceedings of the {Twenty}-first {International} {Joint} {Conference} on {Artificial} {Intelligence}},
	publisher = {AAAI},
	author = {Foster, Mary Ellen and Giuliani, Manuel and Isard, Amy and Matheson, Colin and Oberlander, Jon and Knoll, Alois},
	year = {2009},
}

@article{cangelosi_review_2018,
	title = {A review of abstract concept learning in embodied agents and robots},
	volume = {373},
	issn = {09628436 (ISSN)},
	url = {https://www.scopus.com/inward/record.uri?eid=2-s2.0-85048804487&doi=10.1098%2frstb.2017.0131&partnerID=40&md5=83b68d4eee2d57022bee6f80269dbbbf},
	doi = {10.1098/rstb.2017.0131},
	language = {English},
	number = {1752},
	journal = {Philosophical Transactions of the Royal Society B: Biological Sciences},
	author = {Cangelosi, A. and Stramandinoli, F.},
	year = {2018},
	pages = {6},
}

@inproceedings{sherer_follow_2025,
	address = {Singapore},
	title = {Follow {Me}: {A} {Study} on the {Dynamics} of {Alignment} {Between} {Humans} and {LLM}-{Based} {Social} {Robots}},
	isbn = {978-981-96-3519-1},
	shorttitle = {Follow {Me}},
	doi = {10.1007/978-981-96-3519-1_44},
	language = {en},
	booktitle = {Social {Robotics}},
	publisher = {Springer Nature},
	author = {Sherer, Jeffrey and McPherson, Robbie and Mohanty, Sattwik and Santé, Guilhem and Gandolfi, Greta and Romeo, Marta and Suglia, Alessandro},
	editor = {Palinko, Oskar and Bodenhagen, Leon and Cabibihan, John-John and Fischer, Kerstin and Šabanović, Selma and Winkle, Katie and Behera, Laxmidhar and Ge, Shuzhi Sam and Chrysostomou, Dimitrios and Jiang, Wanyue and He, Hongsheng},
	year = {2025},
	pages = {487--496},
}

@article{brennan_conceptual_1996,
	title = {Conceptual pacts and lexical choice in conversation},
	volume = {22},
	issn = {1939-1285},
	doi = {10.1037/0278-7393.22.6.1482},
	number = {6},
	journal = {Journal of Experimental Psychology: Learning, Memory, and Cognition},
	author = {Brennan, Susan E. and Clark, Herbert H.},
	year = {1996},
	pages = {1482--1493},
}

@article{bobu_learning_2022,
	title = {Learning {Perceptual} {Concepts} by {Bootstrapping} {From} {Human} {Queries}},
	volume = {7},
	copyright = {https://ieeexplore.ieee.org/Xplorehelp/downloads/license-information/IEEE.html},
	issn = {2377-3766},
	url = {https://www.scopus.com/inward/record.uri?eid=2-s2.0-85135735798&doi=10.1109%2fLRA.2022.3196164&partnerID=40&md5=951e31e893493faa5de51fb6c5a409fc},
	doi = {10.1109/LRA.2022.3196164},
	language = {en-US},
	number = {4},
	urldate = {2024-03-29},
	journal = {IEEE ROBOTICS AND AUTOMATION LETTERS},
	author = {Bobu, Andreea and Paxton, Chris and Yang, Wei and Sundaralingam, Balakumar and Chao, Yu-Wei and Cakmak, Maya and Fox, Dieter},
	month = oct,
	year = {2022},
	pages = {11260--11267},
}

@inproceedings{mohapatra_conversational_2023,
	address = {Paris France},
	title = {Conversational {Grounding} in {Multimodal} {Dialog} {Systems}},
	isbn = {979-8-4007-0055-2},
	url = {https://dl.acm.org/doi/10.1145/3577190.3614226},
	doi = {10.1145/3577190.3614226},
	language = {en},
	urldate = {2025-09-12},
	booktitle = {{INTERNATIONAL} {CONFERENCE} {ON} {MULTIMODAL} {INTERACTION}},
	publisher = {ACM},
	author = {Mohapatra, Biswesh},
	month = oct,
	year = {2023},
	pages = {706--710},
}

@article{olivares-alarcosReviewComparisonOntologybased2019a,
    title = {A review and comparison of ontology-based approaches to robot autonomy},
    volume = {34},
    issn = {0269-8889, 1469-8005},
    url = {https://doi.org/10.1017/S0269888919000237},
    doi = {10.1017/S0269888919000237},
    language = {en},
    urldate = {2025-04-02},
    journal = {The Knowledge Engineering Review},
    author = {Olivares-Alarcos, Alberto and Beßler, Daniel and Khamis, Alaa and Goncalves, Paulo and Habib, Maki K. and Bermejo-Alonso, Julita and Barreto, Marcos and Diab, Mohammed and Rosell, Jan and Quintas, João and Olszewska, Joanna and Nakawala, Hirenkumar and Pignaton, Edison and Gyrard, Amelie and Borgo, Stefano and Alenyà, Guillem and Beetz, Michael and Li, Howard},
    month = jan,
    year = {2019},
    pages = {e29},
}

@inproceedings{sauppeDesignPatternsExploring2014,
    address = {Toronto, Ontario, Canada},
    title = {Design patterns for exploring and prototyping human-robot interactions},
    isbn = {978-1-4503-2473-1},
    url = {http://dl.acm.org/citation.cfm?doid=2556288.2557057},
    doi = {10.1145/2556288.2557057},
    language = {en},
    urldate = {2020-11-12},
    booktitle = {Proceedings of the 32nd annual {ACM} conference on {Human} factors in computing systems - {CHI} '14},
    publisher = {ACM Press},
    author = {Sauppé, Allison and Mutlu, Bilge},
    year = {2014},
    pages = {1439--1448},
}

@article{onnaschTaxonomyStructureAnalyze2021a,
    title = {A {Taxonomy} to {Structure} and {Analyze} {Human}–{Robot} {Interaction}},
    volume = {13},
    issn = {1875-4791, 1875-4805},
    url = {https://link.springer.com/10.1007/s12369-020-00666-5},
    doi = {10.1007/s12369-020-00666-5},
    language = {en},
    number = {4},
    urldate = {2026-03-27},
    journal = {International Journal of Social Robotics},
    author = {Onnasch, Linda and Roesler, Eileen},
    month = jul,
    year = {2021},
    pages = {833--849},
}

@inproceedings{jiangMixedInitiativeHumanRobotInteraction2015a,
    address = {Kowloon Tong, Hong Kong},
    title = {Mixed-{Initiative} {Human}-{Robot} {Interaction}: {Definition}, {Taxonomy}, and {Survey}},
    isbn = {978-1-4799-8697-2},
    shorttitle = {Mixed-{Initiative} {Human}-{Robot} {Interaction}},
    url = {http://ieeexplore.ieee.org/document/7379306/},
    doi = {10.1109/SMC.2015.174},
    language = {en},
    urldate = {2026-03-27},
    booktitle = {2015 {IEEE} {International} {Conference} on {Systems}, {Man}, and {Cybernetics}},
    publisher = {IEEE},
    author = {Jiang, Shu and Arkin, Ronald C.},
    month = oct,
    year = {2015},
    pages = {954--961},
}

@inproceedings{holthausCommunicativeRobotSignals2023a,
    address = {Stockholm Sweden},
    title = {Communicative {Robot} {Signals}: {Presenting} a {New} {Typology} for {Human}-{Robot} {Interaction}},
    isbn = {978-1-4503-9964-7},
    shorttitle = {Communicative {Robot} {Signals}},
    url = {https://dl.acm.org/doi/10.1145/3568162.3578631},
    doi = {10.1145/3568162.3578631},
    language = {en},
    urldate = {2026-03-27},
    booktitle = {Proceedings of the 2023 {ACM}/{IEEE} {International} {Conference} on {Human}-{Robot} {Interaction}},
    publisher = {ACM},
    author = {Holthaus, Patrick and Schulz, Trenton and Lakatos, Gabriella and Soma, Rebekka},
    month = mar,
    year = {2023},
    pages = {132--141},
}

@inproceedings{kahnDesignPatternsSociality2008,
    address = {Amsterdam, The Netherlands},
    title = {Design patterns for sociality in human-robot interaction},
    isbn = {978-1-60558-017-3},
    url = {http://portal.acm.org/citation.cfm?doid=1349822.1349836},
    doi = {10.1145/1349822.1349836},
    language = {en},
    urldate = {2020-11-12},
    booktitle = {Proceedings of the 3rd international conference on {Human} robot interaction  - {HRI} '08},
    publisher = {ACM Press},
    author = {Kahn, Peter H. and Freier, Nathan G. and Kanda, Takayuki and Ishiguro, Hiroshi and Ruckert, Jolina H. and Severson, Rachel L. and Kane, Shaun K.},
    year = {2008},
    pages = {97},
}

@inproceedings{10.1145/3411764.3445767,
    address = {Yokohama Japan},
    series = {{CHI} '21},
    title = {Patterns for {Representing} {Knowledge} {Graphs} to {Communicate} {Situational} {Knowledge} of {Service} {Robots}},
    copyright = {All rights reserved},
    isbn = {978-1-4503-8096-6},
    url = {https://dl.acm.org/doi/10.1145/3411764.3445767},
    doi = {10.1145/3411764.3445767},
    language = {en},
    urldate = {2022-05-05},
    booktitle = {Proceedings of the 2021 {CHI} {Conference} on {Human} {Factors} in {Computing} {Systems}},
    publisher = {ACM},
    author = {Zhang, Shengchen and Wang, Zixuan and Chen, Chaoran and Dai, Yi and Ye, Lyumanshan and Sun, Xiaohua},
    month = may,
    year = {2021},
    pages = {1--12},
}

@article{pickeringMechanisticPsychologyDialogue2004,
    title = {Toward a mechanistic psychology of dialogue},
    volume = {27},
    issn = {0140-525X, 1469-1825},
    url = {http://www.journals.cambridge.org/abstract_S0140525X04000056},
    doi = {10.1017/S0140525X04000056},
    language = {en},
    number = {02},
    urldate = {2025-01-05},
    journal = {Behavioral and Brain Sciences},
    author = {Pickering, Martin J. and Garrod, Simon},
    month = apr,
    year = {2004},
}

@article{gastaldonLinguisticAlignmentArtificial2025,
    title = {Linguistic alignment with an artificial agent: {A} commentary and re-analysis},
    volume = {259},
    issn = {00100277},
    shorttitle = {Linguistic alignment with an artificial agent},
    url = {https://linkinghub.elsevier.com/retrieve/pii/S0010027725000393},
    doi = {10.1016/j.cognition.2025.106099},
    language = {en},
    urldate = {2026-03-29},
    journal = {Cognition},
    author = {Gastaldon, Simone and Calignano, Giulia},
    month = jun,
    year = {2025},
    pages = {106099},
}

\end{document}